\pdfoutput=1

\documentclass[11pt]{article}

\usepackage{comment}
\usepackage{acl}
\usepackage{graphicx}
\usepackage{hyperref}
\usepackage{times}
\usepackage{latexsym}
\usepackage{caption}
\usepackage{subcaption}
\usepackage[T1]{fontenc}

\usepackage[utf8]{inputenc}
\usepackage{algorithm}
\usepackage{algpseudocode}
\usepackage{booktabs}
\usepackage{booktabs}
\usepackage{xcolor}
\usepackage{caption}
\usepackage{tabularx}  
\usepackage{microtype}
\definecolor{kmy-color}{rgb}{0.858, 0.188, 0.478}

\title{UNO-DST: Leveraging Unlabelled Data in Zero-Shot Dialogue State Tracking }

\author{Chuang Li$^{1,2}$, Yan Zhang$^{1}$,  Min-Yen Kan$^{1}$, Haizhou Li$^{1,3}$ \\
        $^{1}$National University of Singapore \\
        $^{2}$NUS Graduate School for Integrative Sciences and Engineering\\
        $^{3}$Chinese University of Hong Kong, Shenzhen\\
        \texttt{lichuang@u.nus.edu}\\
        \texttt{\{eleyanz, kanmy, haizhou.li\}@nus.edu.sg}}

\begin{document}
\maketitle

\newcommand{\Vic}[1]{\textcolor{blue}{$_{Vic}$[#1]}}

\begin{abstract}
Previous zero-shot dialogue state tracking (DST) methods only apply transfer learning, ignoring unlabelled data in the target domain.
We transform zero-shot DST into few-shot DST by utilising such unlabelled data via joint and self-training methods. Our method incorporates auxiliary tasks that generate slot types as inverse prompts for main tasks, creating slot values during joint training.  Cycle consistency between these two tasks enables the generation and selection of quality samples in unknown target domains for subsequent fine-tuning. This approach also facilitates automatic label creation,
thereby optimizing the training and fine-tuning of DST models. We demonstrate this method's effectiveness on general language models in zero-shot scenarios, improving average joint goal accuracy by $8\%$ across all domains in MultiWOZ\footnote{Code and data are available at  \href{https://github.com/lichuangnus/UNO-DST}{https://github.com/lichuangnus/UNO-DST}}.

\end{abstract}

\section{Introduction}

\begin{figure}
  \includegraphics[width=0.47\textwidth]{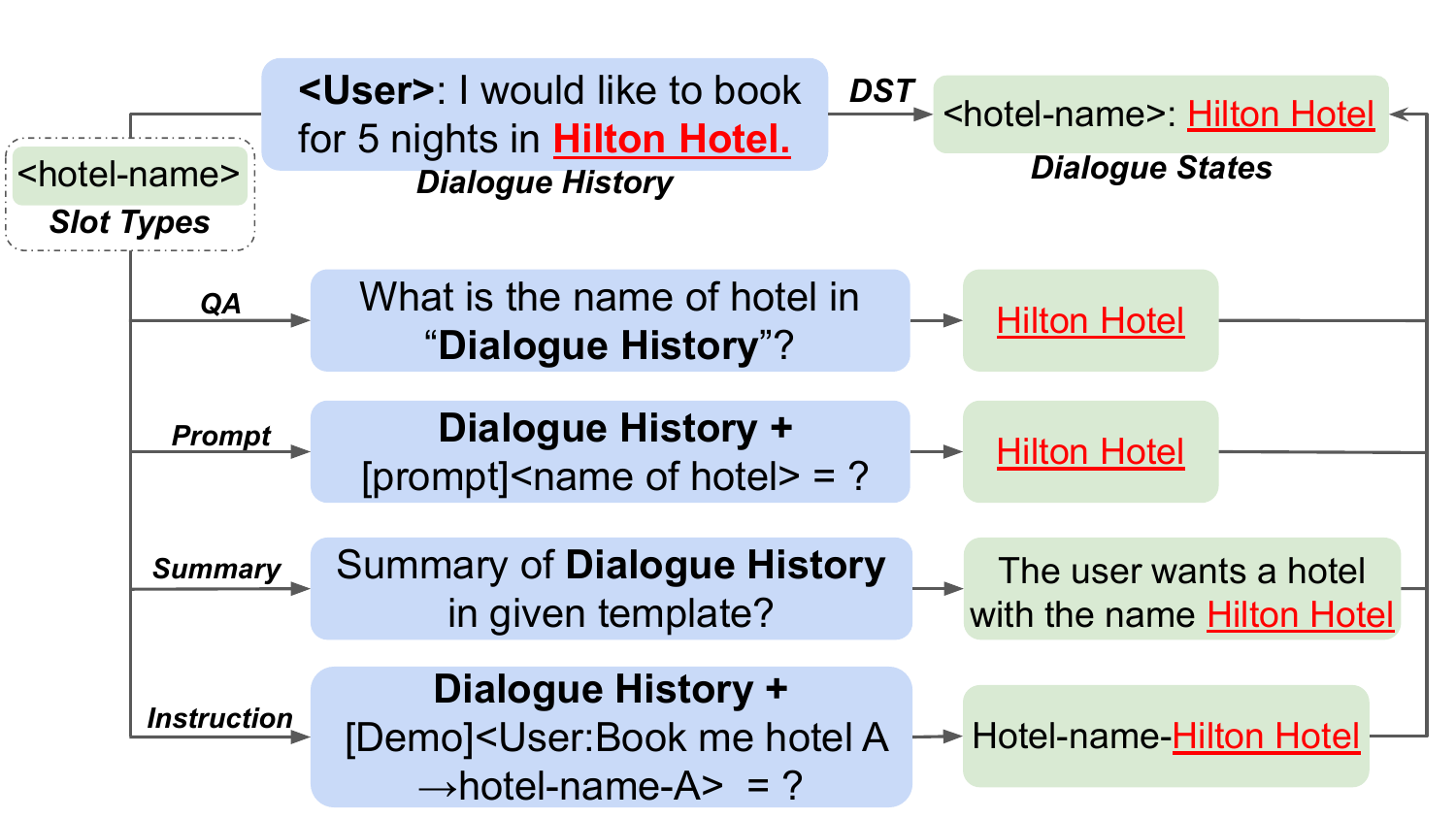}
  \caption{Examples of zero-shot methods in DST.} 
  \label{fig:zero-shot}
\end{figure}

Dialogue state tracking (DST) is a crucial task in understanding users' intentions by extracting the dialogue states from the dialogue history \cite{balaraman_recent_2021}, where a single dialogue state is a pairing of a slot type (e.g.,\textit{<hotel-name>}) and a slot value ({e.g.,\textit{<Hilton hotel>}}), as in Figure~\ref{fig:zero-shot}. Dialogue states are a set of those combinations (e.g.,\textit{<hotel-name: Hilton hotel>}) retrieved by DST models, given dialogue history and slot types. Traditional methods train and evaluate DST models with manually-labelled dialogue states in each domain, which can be costly and time-consuming \cite{wu_improving_2020, simpleTOD}. 
Recently, DST under zero and few-shot settings draw increased attention \cite{lin_leveraging_2021,  hudecek_discovering_2021}. Compared with few-shot methods, zero-shot approaches are more challenging, due to unseen slot types and data scarcity in unknown target domains. 

In both zero and few-shot settings, the majority of existing methods
convert the DST problem into other common problem settings in natural language processing (NLP): for example, Question Answering (QA; \citealp{lin_zero-shot_2021, li-2021-zero}), prompt learning \cite{lee_dialogue_2021}, summarization \cite{DS2-summariztion} and instruction learning \cite{show-don't-tell}. For example in Figure~\ref{fig:zero-shot}, a given slot type (\textit{<hotel-name>})  can be transformed into a QA setting by queries like ``What is the hotel name mentioned in the dialogue history?'' and the slot values can be predicted by a QA model accordingly. 

Transfer learning methods also convert DST tasks to generation ones, more suited for pre-trained language models (LMs; \citealp{BERT}). However, such methods cannot fully leverage the capability of LMs in generation and selection. 
Two main difficulties emerge: 1) the performance of the chosen NLP tasks can be unpredictable for unseen slot types in a new domain due to domain divergence; and 2) existing models are only trained in the known domains, without utilizing any unlabeled data in the new target domain.

\begin{figure}
  \includegraphics[width=0.47\textwidth]{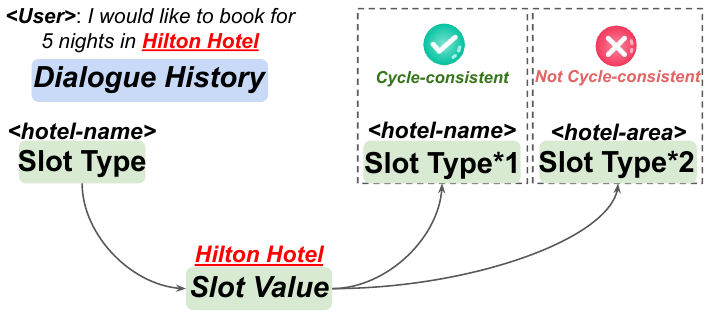}
  \caption{Cycle consistency in DST.} 
  \label{cycle}
\end{figure}

\begin{figure*}

  \centering
  \includegraphics[width=0.98\textwidth]{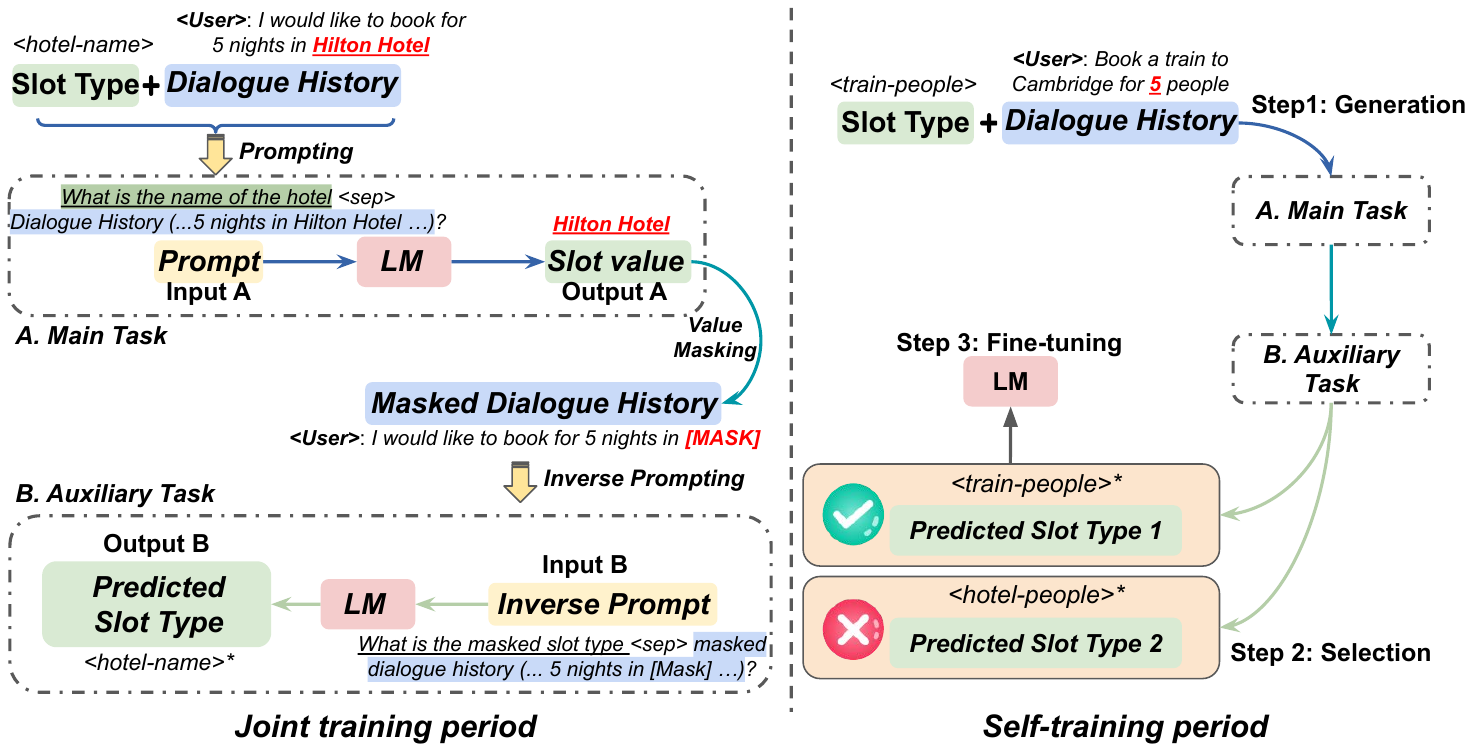}

\caption{Overview of \textbf{UNO-DST} which consists of two periods: 1) joint training for both task A (slot value prediction) and B (slot type prediction), 2) self-training in the target domain. Step 1: Generation of slot values and types; Step 2: Selection of good samples with cycle consistency; Step 3: Fine-turning the LM with selected samples.}


\label{main_methods}
\end{figure*}

This work proposes \textbf{UNO-DST}\footnote{``Uno'', Spanish for ``one'', embodies our proposed strategy in this paper: transitioning from zero to one and subsequently from one to all.}, a method to leverage the \underline{\textbf{un}}labelled data for \underline{\textbf{zero}}-shot \underline{\textbf{DST}} in target domain. Inspired by the popularity of multi-task learning and self-supervised learning \cite{MTsurvey, tsai_self-supervised_2021}, UNO-DST employs a two-step training framework invoking both joint and self-training
(Figure~\ref{main_methods}). 
Aside from the main task of generating slot values, we design an auxiliary task of generating slot types.  We then jointly train both tasks using the labelled training data in source domains. 
For the self-training period, we implement the concept of cycle consistency within our two tasks \cite{CycleGAN2017, dual-prompt}.  That is, a text output from the main task serves as input to the auxiliary task, and the resultant text produced by the auxiliary task should match the original input text (Figure~\ref{cycle}). This process forms a full cycle, ensuring consistent generation and selection of dialogue states from the unlabelled data, which is further used for fine-tuning the model. 
In this way, we convert zero-shot problems into few-shot ones.  Importantly, our framework is model-agnostic which applies to different baseline models.
Our main contributions are as follows:
\begin{itemize}
\setlength\itemsep{0.01cm}
\item To the best of our knowledge, we are the first zero-shot DST work to use unlabelled training data in an unknown target domain;
\item We introduce an auxiliary task to facilitate the training of the main task, the selection of fine-tuning samples, and the generation of unseen or new slot types;
\item We demonstrate our methods with encoder-decoder LMs and large language models (LLMs), showing its effectiveness on two popular DST datasets.  
\end{itemize}

\section{Related Works}

Existing DST methods are generally classified as either 1) full-data  
or 2) low-resource DST.  Despite the method chosen, unlabeled training data in the target domain remains unexploited; in few-shot DST, although pseudo labels can be derived from unlabeled data in the same domain \cite{lee23k_interspeech}, there is a notable absence of research on target domain unlabelled data in zero-shot DST.


\textbf{Full-data DST} are commonly trained with fully annotated multi-domain conversations \cite{tod-bert, simpleTOD}. 
SOTA models focus on DST tasks with well-annotated datasets \cite{data-driven-dst, spoken-dst}. However, the annotation work for data in a new domain can be costly. Hence there is interest in transferring the knowledge of a model from a known domain into an unknown domain and conducting DST tasks in a low-resource setting. 


\textbf{Low-resource DST} uses zero- or few-shot learning in the unknown target domain. Here, the state-of-the-art use a single NLP task to transfer knowledge from the source domains to the unknown target domain \cite{lin_leveraging_2021, DS2-summariztion}. 
While transfer learning tasks achieve good results, each method is task-dependent. Thus, task-independent strategies have been proposed  \cite{ SD-T5, dual-prompt}. 

\textbf{Multi-task Learning} involves simultaneous training of a model on diverse tasks, to boost performance on trained downstream tasks. It also holds promise for enhancements on new tasks \cite{t5, MTsurvey}. However, existing methods typically neglect to assess the consistency across multiple tasks after joint training, while our approach leverages the cycle consistency for selection \cite{CycleGAN2017, wang2023selfconsistency}.

\section{Methodology}
In Figure~\ref{main_methods}, we show an overview of \textbf{UNO-DST} with joint training and self-training periods. Our method includes two tasks: a main task for slot value prediction (\S\ref{main task}) and an auxiliary task for slot type prediction (\S\ref{auxiliary task}). In the joint training period, both tasks are jointly trained in the known source domains (\S\ref{joint section}). In the self-training period, we introduce three steps to generate dialogue states, select good samples, and fine-tune the LM (\S\ref{self-training}). Lastly, we elaborate on the transferability of our strategy with an oracular selection approach (\S\ref{lower/upper bound}). 

\subsection{Task Definition}\label{main task}

The main task for DST is predicting the \textit{<slot-type:slot-value>} pairs with given dialogue history and slot types from a pre-defined slot type list, as shown in Figure~\ref{fig:zero-shot}. 
For each domain, there are seen slot types which appear in other domains (e.g.,\textit{``hotel-name'' and ``restaurant-name''}) or unseen slot types which are unique in the specific domain (e.g.,\textit{``hotel-stars''}). 
The number of these unseen slot types represents the difficulty of zero-shot DST for each domain \cite{SD-T5}.

We denote the dialogue history in a $t$-turn conversations as $C_t = \{c_1, c_2 ... c_t\}$ and slot types $S$ in domain $h$ as $S_h = \{s_1, s_2 ... s_n\}$. For each conversation turn, the main goal is to predict slot values $v'$. Therefore, the input for the $LM$ is combined of dialogue history and slot types, with the output being slot values, as shown in Eq.~\ref{1}. 
\begin{equation}
v'_i = LM(C_t, s_i)
\label{1}
\end{equation}

Compared with methods that select slot values from a constant ontology list using classification models \cite{shi2017convolutional}, we enhance the capability of text-to-text LMs for text generation \cite{ heck2020trippy}. For the case when there are no slot values related to a given slot type, we train the model to output a \textit{``none''} value, indicating that there are no dialogue states from the current conversation turn.

To better utilise the capability of LMs in different tasks, we utilize different prompt functions ``$P(.)$'' to generate the prompt in the correct format. For example, given a slot $s$ and context $c$, the QA prompt $p$ for the DST can be $p^{main} = $\textit{``What is the value of slot $s$ in context $c$?''}. 
We formulate the way of using prompts for the DST main task as: 
\begin{equation}
v'_i = LM(p_i^{main}) = LM(P(s_i, C_t))
\end{equation}

\subsection{Auxiliary task}\label{auxiliary task}
As joint training can improve the accuracy of LMs, we design an auxiliary task to facilitate the training of the main task \cite{MTsurvey, su-multi-2022, dual-prompt}. We propose an auxiliary task to help the model better understand the semantic and context information from the dialogue history in the joint training period and serve as a regulator to check the main task predictions obtained during the self-training period.

We design the auxiliary task as the inverse (converse) prompt of the main task. In opposition to the main task, the auxiliary task thus takes the slot values $v$ as input and generates the slot types $s'$ as outputs, which forms a cycle-consistent loop as a foil to the main task. To make it easier for LMs, we convert the slot values $v$ and dialogue history $C_t$ into a masked dialogue history $C_t^m$ for the model to make better masked predictions, as in Eq.~\ref{5}. The inverse QA prompt $p^{aux}$ is generated as \textit{``What is the masked slot type in context $C_t^m$?''} from inverse prompt function ``$IP(.)$'' (Figure~\ref{main_methods}). We implement the auxiliary task during both the joint training and self-training periods to facilitate slot values generation and selection.
\begin{equation}
s'_i = LM(p_i^{aux}) = LM(IP(v_i, C_t^m))
\label{5}
\end{equation}
\subsection{Joint training with auxiliary tasks}\label{joint section}
We conduct a simple version of joint training with only two tasks: the main and the auxiliary DST tasks. The training samples for the main tasks are created using dialogue history and slot type, while the samples for the auxiliary DST tasks are created by masking the slot values from dialogue history. 


As the auxiliary task is an inverse process of the main task, the model is trained for the same knowledge in a cycle-consistent way. By predicting the masked slot type from the masked dialogue history, the model is familiar with the context and different slot types. With our specially designed auxiliary task, the generation model reuses the existing data for another round of training without the need to increase the amount of training data or model parameters. We formulate the loss function for the main task $L_m$ and auxiliary task $L_a$ as: 
\begin{equation}
L_m= -\sum^n_i{\log p(v'_i|C_t, s_i)}
\end{equation}
\begin{equation}
L_a= -\sum^n_i{\log p(s'_i|C_t^m, v_i)}
\end{equation}
The final loss is a simple average of both. To keep the process simple, we do not add hyperparameters to the model framework. Importantly, as the auxiliary task samples are generated using the inverse prompt of the main task, the ratio of these two tasks mirrors the natural distribution of both tasks throughout the joint training period.
\begin{table}[!t]
\small
\centering
\begin{tabular}{llll}
\toprule
    & \multicolumn{1}{c}{\textbf{MultiWOZ}} & \multicolumn{2}{c}{\textbf{SGD}}  \\
    \cmidrule(lr){2-2} \cmidrule(lr){3-4} 
    & \multicolumn{1}{l}{\textbf{JGA}} & \multicolumn{1}{l}{\textbf{JGA}} & \multicolumn{1}{l}{\textbf{AGA}}  \\ 
\midrule  
    \textbf{Benchmarks} & 25.8 & 27.6 & 58.0\\
    \textbf{T5DST} & 32.4 & NA & NA\\
    \textbf{SD-T5} & 35.6 & NA & NA\\
    \textbf{TransferQA} & 35.8 & 21.3 & 60.8  \\ 

    \textbf{UNO (JT)} & 
    \textbf{36.6} \textcolor{red}{(+0.8)} & 
    \textbf{36.9} \textcolor{red}{(+15)} &
    \textbf{75.9} \textcolor{red}{(+15)} \\
    \textbf{UNO (JT-ST)} & 
    \textbf{40.8} \textcolor{red}{(+5.0)} & 
    \textbf{47.4} \textcolor{red}{(+26)} &
    \textbf{81.8} \textcolor{red}{(+21)} \\
\bottomrule
\end{tabular}
\caption{Average zero-shot JGA and AGA results on MultiWOZ and SGD.  JT/ST stands for joint/self-training and \textcolor{red}{red} figures calculate the performance increase of UNO-DST over TransferQA.}
\label{tab:overall}
\end{table}

\subsection{Self-training with auxiliary tasks} \label{self-training}

Compared with other zero-shot DST models, the key novelty of our strategy is in using the unlabelled training data in the unknown target domain for self-training. Self-training aims to generate pseudo labels and select data samples that further fine-tune the models. 
In the self-training period, we divide the strategy into three steps: termed generation, selection and fine-tuning.

{\bf Step 1 Generation} tests both tasks using the unlabelled training data in the unknown target domain to generate predicted slot values $v'$ and slot types $s'$. Auxiliary tasks in self-training are created by value masking, as shown in Figure~\ref{main_methods}. 
For training samples with slot values that do not directly copy from the original context (such as \textit{``yes/no''} for \textit{``hotel-parking''}), masking the slot value in the original context does not work.  Such samples are omitted in creating the masked dialogue history.

{\bf Step 2 Selection} tests the cycle consistency between main and auxiliary tasks by comparing the predicted slot types $s'$ with the original slot types $s$ in each dialogue turn. A simplified selection process is shown in Figure~\ref{main_methods}. In experiments, only the conversations with fully correct slot types are selected as good samples like joint goal accuracy settings, aiming to reduce the selection error. 

{\bf Step 3 fine-tunes} the model $LM(.)$ with selected samples and predicted slot values $v'$.  This completes the conversion of zero-shot DST into few-shot DST, helping the model adapt to unknown domains without increasing data annotation and model parameters. For LLMs that are difficult to fine-tune, we propose other solutions (\S~\ref{Connection to LM}).


\subsection{Oracular selection for zero-shot DST} \label{lower/upper bound}

Even though there are many studies working on zero-shot DST, to the best of our knowledge there are no common methods to identify the 
the peak performance that each model can potentially achieve (oracular performance).
Here, we discuss our proposed algorithm with respect to the oracular selection, aiming to benchmark our method against oracular results for each model as an upper bound. 

According to our self-training methods (\S\ref{self-training}), zero-shot DST can always be converted into a few-shot DST by selecting good samples with self-generated slot values for fine-tuning. Since cycle consistency cannot ensure $100\%$ correct data selection, oracular performance comes when we select only the correct self-generated samples and use them for fine-tuning. We define such performance as the upper bound for the zero-shot DST model. 
\begin{table*}
\centering
\small
	\begin{center}
	\end{center}
 \setlength{\tabcolsep}{2.1pt}
	\begin{tabular}{  p{1.9cm}p{1.8cm}p{1.8cm}p{1.8cm}p{1.9cm}p{1.8cm}p{1.9cm}p{1.6cm}} 
	\toprule
     \textbf{Model} & \textbf{Checkpoint}  &\textbf{Attraction} & \textbf{Hotel} & \textbf{Restaurant} & \textbf{Taxi} & \textbf{Train} & \textbf{Average} \\ \midrule
    SD-T5 & t5-small &33.9 & 19.9 & 20.8 & 66.3 & 37.0 & 35.6 \\
    
    
    TransferQA & t5-large & 33.9& 22.7 & 26.3& 61.9&36.7&35.8 \\
    \midrule
   T5DST\dag & t5-small  & 30.5&	19.4&	20.4&	\textbf{66.3}&	25.6&	32.4\\\midrule
    UNO (JT) & t5-small & 33.5 &	21.0 	& 22.4& 65.2& 38.7 & 36.2 \\
    UNO (JT-ST)& t5-small & \textbf{36.1} \textcolor{red}{(+5.6)}	& 23.0		& 24.0& 65.0& 48.0	& 39.2 \\
    \midrule
    UNO (JT) & t5-QA & 32.9 &	22.9  	& 29.5& 66.0 \textcolor{white}{-0.3}&	31.7 & 36.6 \\
    UNO (JT-ST) & t5-QA & 33.1&\textbf{25.7} \textcolor{red}{(+6.3)}		&\textbf{31.0} \textcolor{red}{(+10.6)}&65.5	&\textbf{48.9} \textcolor{red}{(+23.3)}&\textbf{40.8} \textcolor{red}{(+8.4)}
     \\
     \bottomrule
    
	\end{tabular}
	
\caption{Zero-shot JGA results with different LM checkpoints. The lower/upper bound and best results for each domain are shown in bold. JT and  ST stand for the results after joint- and self-training. \dag\ shows results of our replicated T5DST model, and \textcolor{red}{red} figures give the performance gap compared to \dag.}
\label{tab:zero-shot}
\end{table*}
\begin{table}[t!]
\small
\centering
\setlength{\tabcolsep}{2pt}
	\begin{center}
	\end{center}
	\begin{tabular}{lrrrrrc}
	\toprule
      & \multicolumn{2}{c}{\textbf{TransferQA}} & \multicolumn{2}{c}{\textbf{UNO (JT)}} &\multicolumn{2}{c}{\textbf{UNO (JT-ST)}} \\ 
    \cmidrule(lr){2-3} \cmidrule(lr){4-5} \cmidrule(lr){6-7} 
    \textbf{Domains} &  \textbf{JGA} & \textbf{AGA} & \textbf{JGA} & \textbf{AGA} & \textbf{JGA} & \textbf{AGA} \\ \toprule
        {Flights}  & 03.6 & 42.9 & \textbf{26.4} & \textbf{75.1} & 25.3 & 72.7 \\
       {RideSharing} & 31.2 & 61.7 & 33.3 & 64.3 & \textbf{73.5} & \textbf{89.8} \\
       {Homes}  & \textbf{31.7} & \textbf{80.6} & 16.8 & 77.6 & 17.9 & 76.3 \\
        {Events}  & 15.6 & 56.8 & 11.5 & 58.0 & \textbf{23.1} & \textbf{71.6} \\
        {Movies}  & 24.0 & 56.2 & 35.5 & \textbf{86.7} & \textbf{52.6} & \textbf{86.7} \\
        {Services}  & 37.2 & 75.6 & 75.1 & 92.1 & \textbf{77.2} & \textbf{92.4} \\
        {Travel}  & 14.0 & 24.2 & 55.2 & 76.7 & \textbf{56.4} & \textbf{77.8} \\
        {Weather}  & 40.3 & 59.4 & 93.8 & 98.0 & \textbf{94.3} & \textbf{98.5} \\
        {Hotels}  & 13.5 & 60.1 & 44.8 & 85.6 & \textbf{75.9} & \textbf{94.6} \\
       {RentalCars}  & \textbf{10.8} & 73.8 & 7.5 & 72.9 & 05.4 & \textbf{79.4} \\
        {Restaurants} & 16.3 & 68.9 & 31.8 & 74.7 & \textbf{35.9} & \textbf{78.5} \\
        {Media}  & 30.2 & 67.5 & 37.0 & 69.7 & \textbf{60.0} & \textbf{89.2} \\
        {Music} & 08.9 & \textbf{62.4} & 11.6 & 54.9 & \textbf{19.1} & 55.5 \\
        \textbf{Average}  & 21.3 & 60.8 & 36.9 & 75.9 & \textbf{47.4} & \textbf{81.8} \\\bottomrule
    \end{tabular}
\caption{Zero-shot JGA and AGA results for domains in SGD dataset. \textbf{Bold} shows the best results and JT/ST stands for joint/self-training.}
\label{tab:SGD-seen}
\end{table}

\section{Experiments and Datasets}


\textbf{Dataset.} We train and test our model on both MultiWOZ~2.1~\cite{multi-woz} and the Schema-Guided Dialogue~(SGD; \citealp{SGD}). MultiWOZ and SGD have dialogues distributed in both training and testing distributions over 7, 13 domains, representing 7K, 16K training examples in English, respectively.  We use standard means for data pre-processing \cite{multi-woz} and follow the MultiWOZ leave-one-out settings for zero-shot training and testing in both datasets \cite{wu_transferable_2019, SGD}.


\textbf{Evaluation metrics.} The primary metric for DST evaluation is joint goal accuracy (JGA), which compares the set of generated predicted values with the set of ground truth ones after each conversation turn and average goal accuracy (AGA) calculates the JGA only for active slot types as in SGD dataset \cite{second-dst-cha, SGD}.

\textbf{Baselines and experiment setup.}
We use T5 \cite{t5} as our baseline model.  For a fair analysis, we also compare our results with previous DST benchmarks: TRADE \cite{wu_transferable_2019} and the SGD baseline \cite{SGD}, and current SOTA models: T5DST \cite{lin_leveraging_2021}, TransferQA \cite{lin_zero-shot_2021} and SD-T5 \cite{SD-T5}. We adopt the cross-domain settings \cite{wu_transferable_2019} for both datasets, experimenting on two checkpoints, ``t5-small''\footnote{\href{https://huggingface.co/t5-small}{https://huggingface.co/t5-small}}
and ``t5-QA''\footnote{\href{https://github.com/facebookresearch/Zero-Shot-DST/}{https://github.com/facebookresearch/Zero-Shot-DST}}. The unlabelled training data in the target domain will be used for self-training and testing data in target domain is only used for final testing. 

We select QA as the main task in our framework for its popularity and test it through different checkpoints holding parameters fixed. We adopt the open-source ``t5-small'' \cite{t5} with 60M parameters as our baseline and train using AdamW with a learning rate of 0.0001, batch size 8 for 1 epoch (zero-shot setting) and 3 epochs (fine-tuning setting) on single GeForce RTX3090.


\section{Results}

Table~\ref{tab:overall} analyzes the zero-shot results of our UNO-DST and the 
 baselines, including the state-of-the-art (SOTA) TransferQA model, across both datasets. 
Our model surpasses all baselines for both joint and self-training phases, inclusive of TransferQA using ``t5-large''. 
Detailed outcomes for each domain and specific training periods across each dataset are presented in Tables~\ref{tab:zero-shot} and~\ref{tab:SGD-seen}.

\subsection{Joint training results}
For the joint training period in the MultiWOZ dataset (Table~\ref{tab:zero-shot}), we are using the same model and prompt as T5DST. UNO-DST shows an increase of more than $4\%$ for JGA across all checkpoints. For SGD (Table~\ref{tab:SGD-seen}), our joint training period increases the previous baseline by even larger margins of $15.6\%$ in JGA ($15.1\%$ in AGA). The joint training period is critical as it prepares a model for self-training.
Table~\ref{tab:zero-shot} shows that using different model checkpoints is also critical even when using the same model architecture and parameter size. The model performs best when we
follow the prompt format in each baseline. 
\begin{table}
\small
\centering
\setlength{\tabcolsep}{2pt}
	\begin{center}
	\end{center}
	\begin{tabular}{cccccccc}
	\toprule
    \textbf{Round}  &\textbf{Att.} & \textbf{Hotel} & \textbf{Res.} & \textbf{Taxi} & \textbf{Train} & \textbf{Avg}& \textbf{Gain} {$\uparrow$} \\ \midrule
    0  & 32.86 &	22.91  	& 29.47& 66.00 &	31.68 & 36.58 \\ 
    1 & 33.09&{25.66}&{30.99}&65.48	&{48.90}&{40.82}&\textcolor{red}{(+4.24)} \\
    2 & 35.53&	27.22&	31.44&	64.71&	54.60&	42.70&\textcolor{red}{(+1.88)} \\
    3 & 36.62&	27.09&	31.14&	65.48&	53.31&	42.73&\textcolor{red}{(+0.03)} \\ 
    \bottomrule
    \end{tabular}
\caption{JGA for multiple rounds of self-training on MultiWOZ.  Absolute gains indicated in \textcolor{red}{red}.}
\label{tab:Self-training}
\end{table}
\subsection{Self-training results}\label{self results}
For the MultiWOZ dataset (Table~\ref{tab:zero-shot}), self-training further improves average JGA by $3.09\%$ after the joint training period. Compared with the baseline, the best performance increases by $8.38\%$ in JGA. 
In SGD dataset (Table~\ref{tab:SGD-seen}), self-training improves the average JGA and AGA in 12 out of 13 domains by an average of $10.5\%$ and $5.9\%$ compared with the joint-training alone, and over $26\%$ and $21\%$ compared to the baseline.

The success of self-training proves the possibility of using pseudo labels generated from zero-shot DST models to bootstrap performance. However, carefully selecting good samples to fine-tune the model is challenging because not all the domains benefit from the self-training process. For example, the result for the \textit{``Taxi''} domain in MultiWOZ and \textit{``Flights''} domain in SGD decreased after self-training. We examine the rationale behind the gains obtained through self-training, which is associated with the gap between joint training and oracular results, as further discussed in \S~\ref{discussion}.



As shown in Table~\ref{tab:Self-training}, as we lengthen self-training from a single round to multiple rounds, our framework's performance continues to improve. However, the performance gap between results from different rounds shows diminishing returns, signalling a plateau. The best result with UNO-DST comes when adding more variation is insignificant and so we stop the training when the margin is below $0.1$ in JGA. 
Future work is required to systematically study this strategy over multi-round self-training.

\section{Discussion} \label{discussion}
\subsection{Oracular selection}
\begin{figure}[!t]
	\centering
	\includegraphics[width=7.5cm]{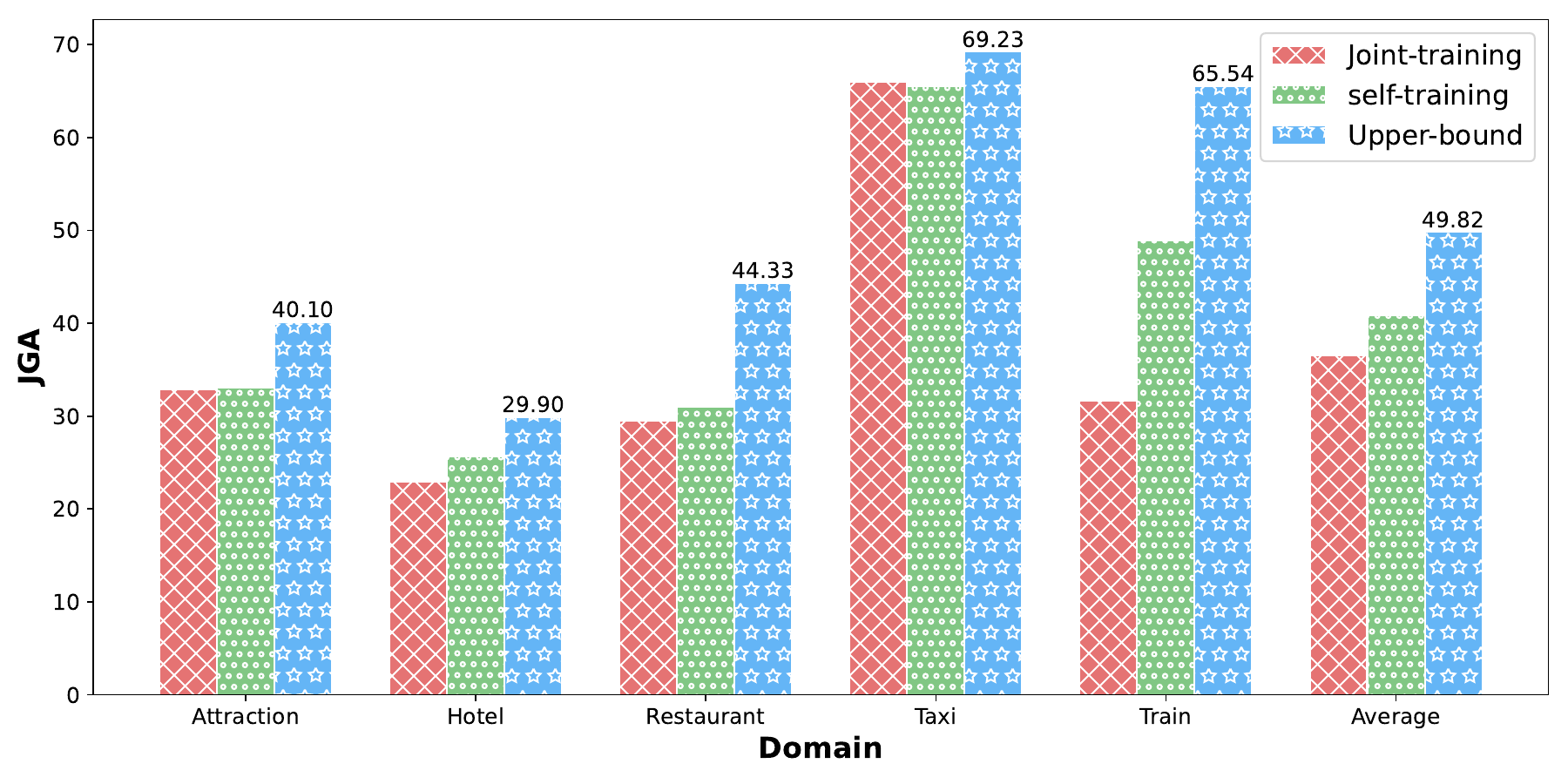}
    \caption{Gains by joint and self-training stages of UNO-DST on the ``t5-QA'' checkpoint.  We show the results of oracular selection (Upper-bound) in each domain for relative comparison. }
    \label{margin}
\end{figure}

For the oracular calculation, we select only the $100\%$ correct samples from the zero-shot predictions and use them for fine-tuning. In Figure~\ref{margin}, we visualise the gains of joint training and self-training alongside our upper bound. While efficacy differs from domain to domain, an important observation is that when the margin between the upper bound (blue columns) and joint training (red columns) is large, the model has a larger gain from self-training, as in \textit{``Train''} domain. In contrast, for \textit{``Taxi''} domain, the influence of self-training is weak (cf \S~\ref{self results}~Self-training results). Utilizing upper bounds calculations enables us to swiftly evaluate whether a domain or model is apt for the self-training period.
In other words, a larger margin between joint training and the upper bound yields a larger potential improvement that the model can achieve with fine-tuning or self-training strategy. 

\subsection{Unseen slot type prediction}
\begin{figure}[!t]
	\centering
	\includegraphics[width=0.47\textwidth]{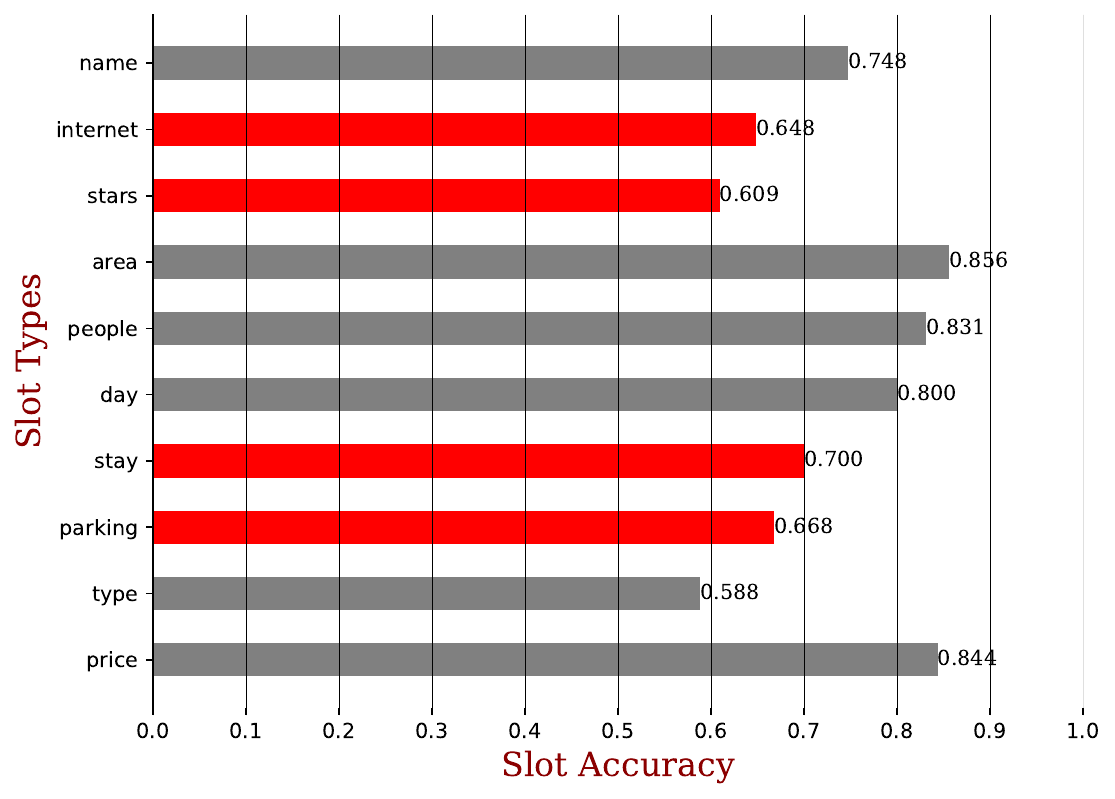}
    \caption{Slot Accuracy for (grey) seen and (red) unseen slot types in the hotel domain.}
    \label{hotel-unseen}
\end{figure}

For each domain, there are seen slot types which appear in other domains (e.g.,\textit{``hotel-name'' and ``restaurant-name''}) or unseen slot types which are unique in the specific domain (e.g.,\textit{``hotel-stars''}). The ratio of the occurrences of these slot types represents the difficulty of zero-shot DST for each domain \cite{SD-T5}.
As shown in table  \ref{slot type a}, the original setting for the MultiWoz dataset has 30 given slot types. However, not all of them appear in every domain. For some certain domains, like the ``hotel'' domain, there are 4 unique slot types which do not appear in other domains, including ``stars'', ``internet'', ``stay'' and ``parking''.  In Figure  \ref{hotel-unseen}, we show the slot accuracy for the hotel domain. It shows that generally, the unseen slot types will perform worse than the seen slot types \cite{SD-T5}. Prediction for those slot types in zero-shot cross-domain settings can be challenging as there is no further information from the other source domains. In addition, half of the unseen slot types in the ``hotel'' domain are related to ``yes/no'' slot values, whereas in our joint training settings in \S~\ref{joint section}, we skip the masking of those ``yes/no'' values from the context and the model is less trained compared with other slot types. We hope that future works will improve on ``yes'' or ``no'' value prediction. 

\begin{table}[!t]
\small

\begin{tabular}{ l } 
 \toprule{\textbf{All Given Slot Types in MultiWOZ  2.1}}\\\midrule
 {$area^{123}, arrive by^{45}, day^{235}, departure^{45},$} 
 \\{$destination^{45}, food^3, internet^2, leave^{45},$}
 \\{$ name^{123}, people^{235}, parking^2, price^{23}, stars^2,$}
 \\{$ stay^2, time^3, type^{12}$}
 \\\midrule {\textbf{Seen Slot Types in Hotel Domain}} 
 \\\midrule
{$area^{123},day^{235},name^{123}, people^{235},price^{23},type^{12}$} 
 \\\midrule {\textbf{Unseen Slot Types in Hotel Domain}} 
 \\\midrule {$internet^2, parking^2,stars^2,stay^2$}
\\\bottomrule
\end{tabular}
\caption{Seen and unseen slot types in hotel domain. The superscript on each slot type indicates the domain information from: (1:attraction, 2:hotel, 3:restaurant, 4:taxi, 5:train)}
\label{slot type a}
\end{table}

\subsection{New slot type generation}\label{new_slot}
All the existing zero-shot DST methods require given slot types in generating the slot values for both source and target domains and our model also follows the same experiment settings (cf \S~\ref{main task}). 
However,  our model can also self-generate reasonable slot types either in or beyond the 30 given slot types with our designed auxiliary task. To self-generate new slot types, we do a case study on the MultiWoZ \textit{``train''} domain and perform random word masking for all the dialogue history, inputting those randomly masked dialogue histories to the auxiliary task for slot type predictions, as shown in Figure~\ref{3b}.
In Table~\ref{slot type b}, we show some valid new slot types generated by our auxiliary tasks with dialogue history. For example, \textit{``asking for the ticket price''} in \textit{``train''} domain and \textit{``asking for parking information''} in \textit{``Restaurant''} domain are reasonable new slot types, which can also be included. 



\subsection{DST without pre-defined slot types}
\begin{figure}[!t]
     \centering
     \begin{subfigure}{0.48\textwidth}
         \centering
         \includegraphics[width=\textwidth]{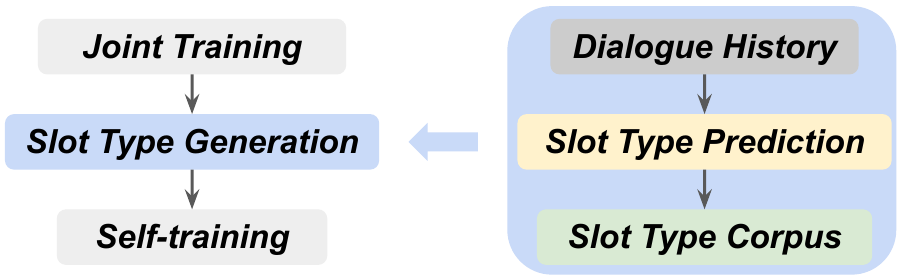}
         \caption{Overall system design for zero-shot DST w/o slot types}
         \label{3a}
     \end{subfigure}
     \hfill
     \begin{subfigure}{0.48\textwidth}
         \centering
         \includegraphics[width=\textwidth]{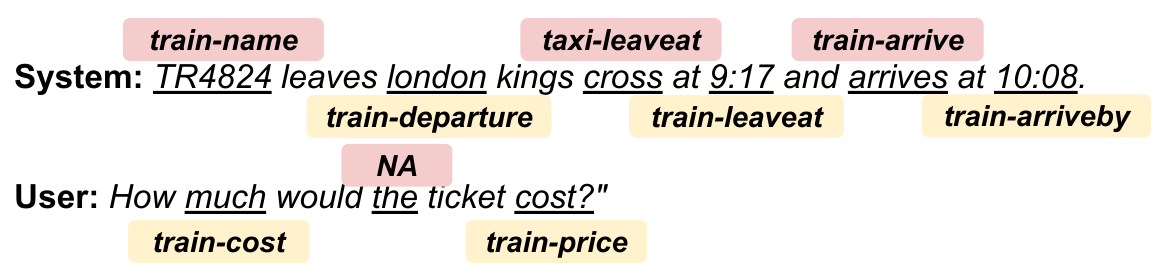}
         \caption{Slot type prediction with randomly masked tokens}
         \label{3b}
     \end{subfigure}
     \hfill
     \begin{subfigure}{0.48\textwidth}
         \centering
         \includegraphics[width=\textwidth]{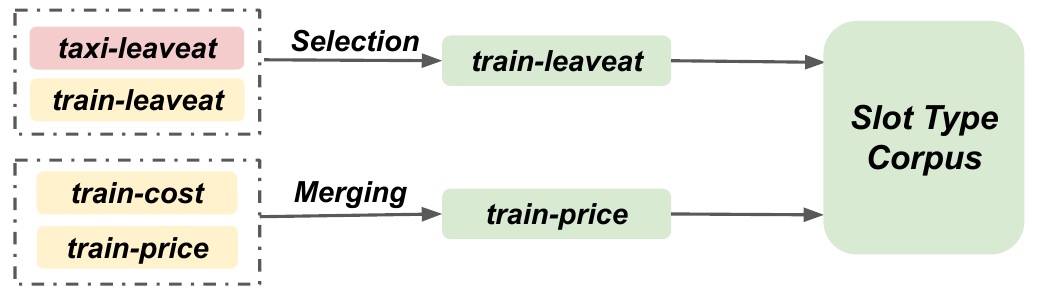}
         \caption{Building slot type corpus with merging and selection}
         \label{3c}
     \end{subfigure}
        \caption{Zero-shot DST without pre-defined slot types}
        \label{fig:three graphs}
\end{figure}

As discussed in the previous section, the auxiliary task can facilitate the generation of new slot types beyond the pre-defined ones in MultiWOZ. Besides adding more slot types to the given slot type list, we believe that our proposed model can conduct DST tasks in an unknown target domain without any given slot types and we describe the proposal of zero-shot DST without slot types in Figure \ref{fig:three graphs}.

In order to eliminate the use of pre-defined slot types, we add a slot type generation period between joint and self-training, which identifies and select domain-relevant slot type corpus.
Similar to the process proposed by  \citeauthor{hudecek_discovering_2021} (\citeyear{hudecek_discovering_2021}), we can first use our auxiliary task to generate potential slot types based on random masked dialogue history, as shown in Figure \ref{3b}. The generated text may contain domain-irrelevant or similar slot types and we propose a weak selection and merging of task-relevant and similar slot types for slot type corpus \cite{hudecek_discovering_2021}. Secondly, those generated slot-type corpus can be used for self-training in the unknown target domain, as discussed in \S~\ref{self-training}. 

During our testing, our auxiliary task can generate predictions including all 6 given slot types in the \textit{``train''} domain (Table~\ref{new_slot}), as well as valid slot types in other domains, which demonstrates the potential of future zero-shot methods without pre-defined slot types. We look forward to future works for zero-shot DST without any labelled data in slot types and values in unknown target domains.

\section{UNO-DST with ChatGPT} \label{Connection to LM}

While earlier sections analyse the effectiveness of our methods on LMs like ``T5'', this section focuses on the potential application of UNO-DST with large language models (LLMs), such as \citealp{GPT3} and \citealp{LLAMA}.
Specifically, we examine UNO-DST for zero-shot DST using OpenAI's ChatGPT\footnote{\href{https://chatgpt.openai.com}{https://chatgpt.openai.com}} as the backbone LLM, an LLM which has been adopted as a language tool for information extraction with strong capabilities, even without specific training or fine-tuning. 

Our study will test the efficacy of our self-training strategy in \textbf{UNO-DST} on ChatGPT, including the web interface for conversational approach (\S\ref{conversations}) and the API\footnote{ ChatGPT API model: {\href{https://openai.com/product}{gpt-3.5-turbo-0301}}} for larger testing corpus using in-context learning (ICL; \citealp{ICL-DST}; \S\ref{icl}).


\subsection{Conversational approaches}\label{conversations}

\begin{figure*}[!t]
     \centering
     \begin{subfigure}[b]{0.49\textwidth}
         \centering
         \includegraphics[width=\textwidth]{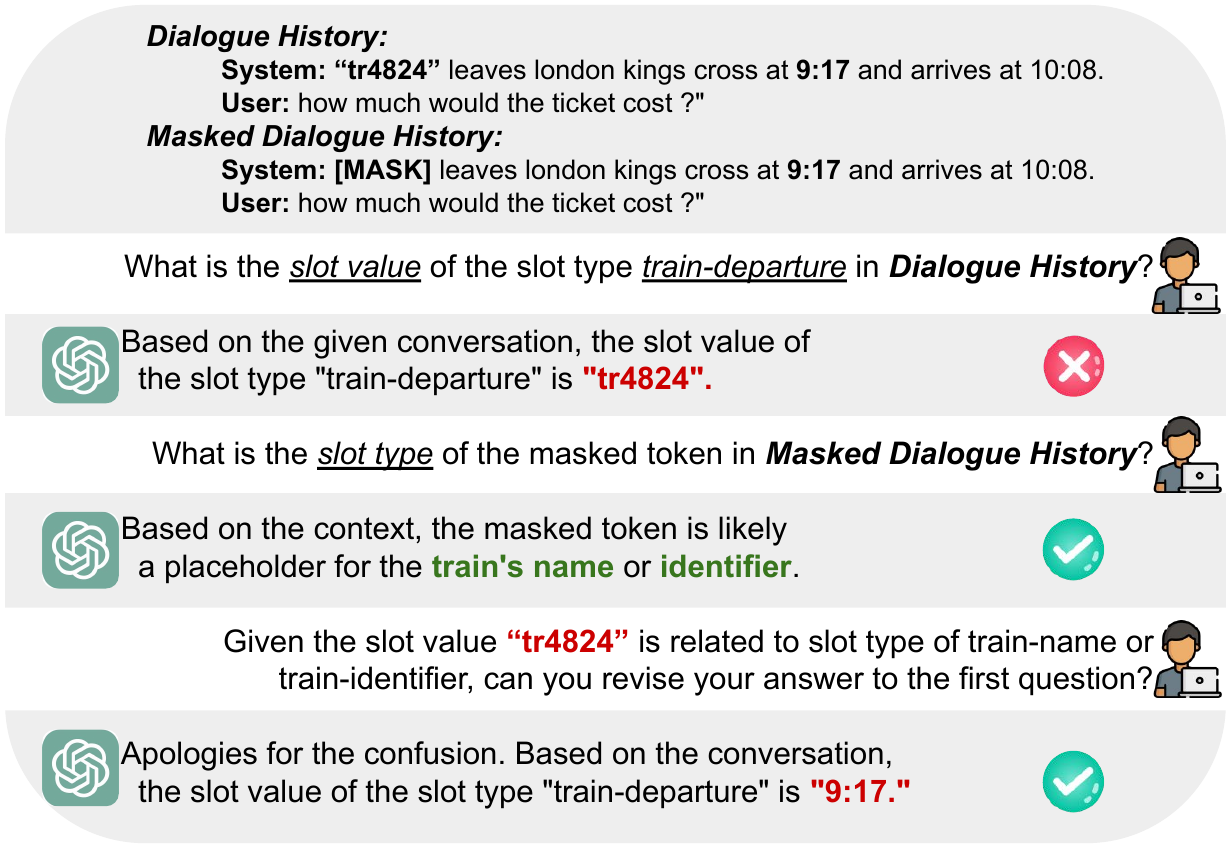}
         \caption{Correction of the wrong prediction for ``train-departure''.}
         \label{2a}
     \end{subfigure}
     \hfill
     \begin{subfigure}[b]{0.49\textwidth}
         \centering
         \includegraphics[width=\textwidth]{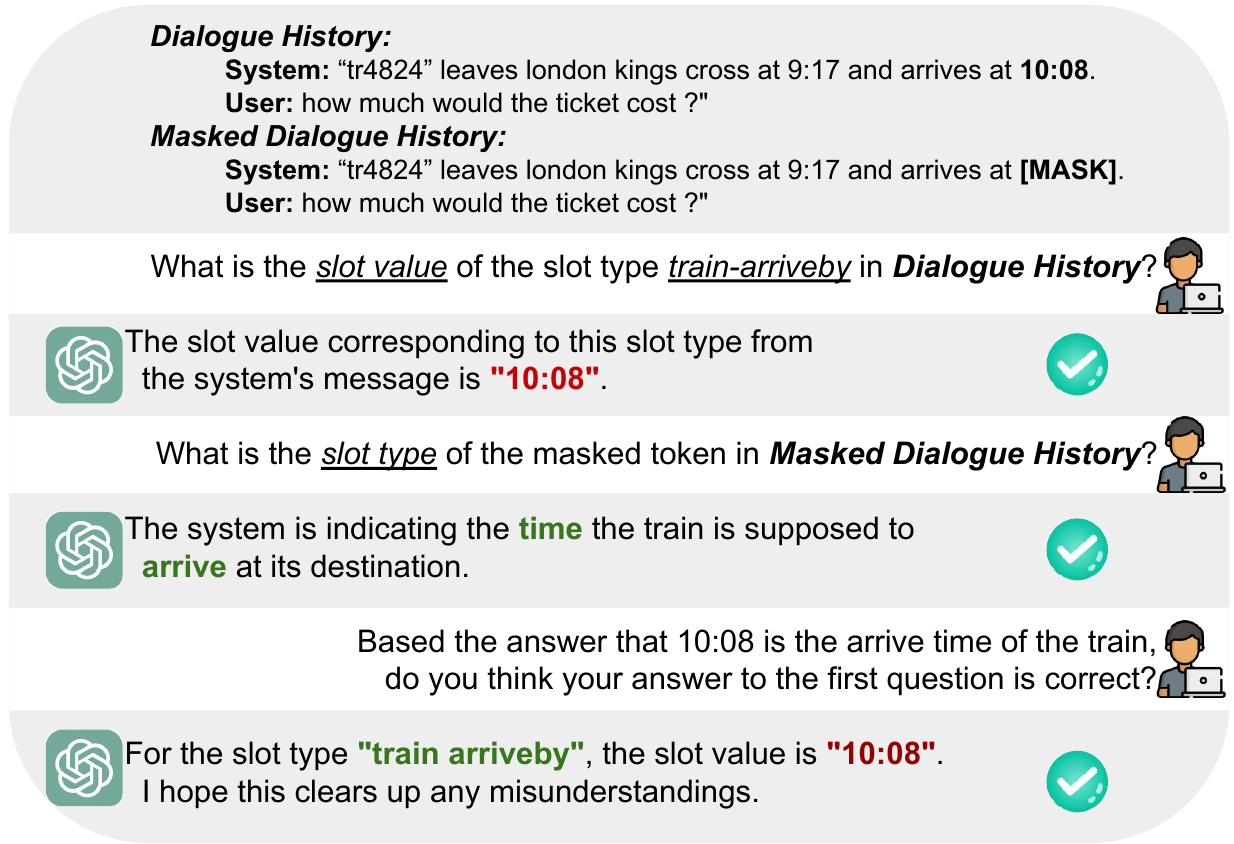}
         \caption{Confirmation of the correct prediction for ``train-arriveby''.}
         \label{2b}
     \end{subfigure}
        \caption{Case studies for conversations with ChatGPT for zero-shot DST. ({\color{red}{Red}}: Slot Values; {\color{green}Green}: Slot Types)}
        \label{fig:two}
\end{figure*}

\textbf{Implementation.}
We skip the joint training for ChatGPT and use conversations as an inference approach. As shown in Figure \ref{fig:two}, we implement main and auxiliary tasks with conversations asking for slot values or types. The selection and fine-tuning steps in the self-training strategy have been converted into the correction or confirmation step, where we provide the slot value and type predictions from the previous two questions to ChatGPT and ask whether it needs to revise the slot value to the main task. We consider the revised response from ChatGPT as the final answer to our main task. We manually examine all the responses generated and give examples of their performance. 

\begin{table}
\small
\centering
\begin{tabular}{ l } 
\\\toprule{\textbf{All Generated Slot Types in Train Domain}}
\\\midrule {$people^5, day^5, destination^5, departure^5, leave^5, $}\\{$arrive^5, price^5, type^5, time^5, area^5, name^5$}
 \\\midrule {\textbf{Valid New Slot Types} } 
 \\\midrule
{${price}^{15}, day^{1}, parking^{3}, name^5$} 
 \\\midrule
 {\textbf{Dialogue Example [PMUL1359] $(price^5, name^5)$}} \\\midrule
{\textbf{System:}\textit{``Okay, tr6572 departs at 05:29.''}}\\
{\textbf{User:}\textit{``What is the price?''}}\\\midrule
 {\textbf{Dialogue Example [PMUL3027] $(parking^3)$}} \\\midrule
{\textbf{System:}\textit{``I have 2 Turkish restaurants in the centre?''}}\\
{\textbf{User:}\textit{``Do they offer free parking?''}}\\\midrule
 {\textbf{Dialogue Example [PMUL1118] $(day^1)$}} \\\midrule
{\textbf{User:}\textit{``I am in Cambridge for the week and want to}}\\
 \hspace{0.9cm}\textit{{know what museums you guys have there.''}}\\\bottomrule
\end{tabular}

\caption{Newly-generated slot types with examples. The superscript on each slot type indicates the domain information from: (1:attraction, 2:hotel, 3:restaurant, 4:taxi, 5:train)}
\label{slot type b}

\label{slot type}
\end{table}

\textbf{Results and discussion.} We show two cases of predictions made by ChatGPT using the same dialogue history in Figure \ref{2a} and \ref{2b}. 
In Figure \ref{2a}, ChatGPT first made a wrong main task prediction, followed by a correct but not consistent auxiliary task prediction. When we provide all historical information to ChatGPT and ask for a revised main task prediction, it realised that the slot value from the first prediction is not consistent with the original slot type and it self-corrected its wrong prediction. 
In Figure \ref{2b}, we show another case of correct predictions which happens for the majority of the conversations. When ChatGPT is able to make correct predictions for both the main and auxiliary tasks, it confirms the correct predictions for the final question based on cycle consistency. 

We illustrate how the cycle consistency between the main and auxiliary tasks aids ChatGPT in rectifying incorrect answers or confirming correct answers \cite{CycleGAN2017}. The strategy applies to the free accessible web interface of ChatGPT and is easy to implement. 
However, as conversations are difficult to quantify and evaluate, we only qualitatively show the results and encourage future research to explore all the potential implementations of cycle consistency in LLMs \cite{wang2023selfconsistency}.

\subsection{In-context Learning}\label{icl}

\textbf{Implementation.}
Following the settings in \S~\ref{conversations}, we skip the joint training period and apply our strategy to improve ChatGPT by ICL, providing ICL instructions and ICL examples in the ICL prompt. Specifically, we compare the JGA results between two different ICL prompts using the same instructions but different ICL examples: 1) ICL examples are composed of both dialogue history and dialogue states in the source domains because no dialogue states in the target domain are available for zero-shot DST settings and 2) ICL examples with dialogue history in the target domain and dialogue states generated using the conversational approach in \S~\ref{conversations}. To illustrate, without cycle consistency, the ICL prompt is originally built by DST examples from the source domains and it is hard for the LLM to test on the target domain. After selecting good samples with the strategy discussed in \S~\ref{conversations}, the ICL prompt can be updated with examples in the target domain. We conduct small-scale experiments $3$ times with the \textit{``train''} domain in the MultiWOZ dataset by randomly sampling $100$ conversations and inference with ICL prompts using different ICL prompts, evaluated by JGA.


\textbf{Results and discussion.}
The resulting average JGA for the original source domain ICL prompt is $34.92\%$ while the JGA for the selected target domain ICL prompt is $54.18\%$.  Our self-training strategy works very well, serving the LLM to select valuable in-domain examples for the ICL prompt, improving the zero-shot DST performance by a large $19.25\% $ margin. By manually examining the generated dialogue states, the ICL prompt modified with our strategy performs better, especially for conversations with longer dialogue turns and more slot types.
Our self-training strategy demonstrates its capability in generating and selecting dialogue state samples in LLMs which can further improve the performance of zero-shot DST using an ICL prompt. However, due to the scope of this paper, we only test the application of our \textbf{UNO-DST} on ChatGPT with a small data corpus.

In summary, this section extends the applicability of the \textbf{UNO-DST} strategy to LLMs, assessing its efficacy in both conversational approaches and ICL. Conversational methods offer a straightforward mechanism for rectifying in-discussion errors, whereas ICL, leveraging APIs or LLM inferences, facilitates handling larger data corpora. Besides testing the cycle consistency strategy in well-suited DST tasks, additional work is required to extend it to other LLMs or more general NLP problems. 

\section{Conclusion}
We propose a novel approach to convert the zero-shot DST into a few-shot setting by generating and selecting quality dialogue states from unlabeled data in the target domain through joint and self-training periods. We introduce and demonstrate how our proposed auxiliary task, which generates slot types as the inverse prompt for the main task which generates slot values, serves the whole model for 1) better accuracy of the main task in joint training 2) quality data selection in the self-training period 3) new slot types generation beyond the given slot type list and 4) upgrading to LLMs. 

Our proposed strategies of \textbf{UNO-DST} are task-independent, which can be extended to other prompt formats and generalised to LLMs. We look forward to future works that engage additional auxiliary tasks which target new datasets and apply zero-shot DST, even where no slot types are given. 

\section*{Limitations}\label{limitations}
This work has 3 limitations: 1) due to the limitation of computational resources, we only conduct experiments on small encoder-decoder LMs, which is ``t5-small'' and simple NLP tasks, which is ``QA''. Our future works will include more NLP tasks with different LMs to systematically test the performance of our proposed models. 2) Our reported self-training results are only for a single round of self-training because we could not find a way to continuously increase the performance of self-training. 
Our future plan seeks to improve and examine the best criteria for self-training using an early-stopping strategy. 3) The experimental settings for ChatGPT can be improved in three aspects: a) a larger data corpus can be applied with better instruction prompts in order to limit ChatGPT in generating more accurate values, b) an open-source LLM \cite{LLAMA} can be applied to better evaluate and replicate the results of the experiment, and c) a full self-training strategy including generation, selection, and fine-tuning can be tested to demonstrate the best performance with LLMs.

\section*{Ethical Concerns}
Our self- and joint-training tunes models to amplify signals from the original dataset.  While this strategy does work well in our experiments, if the dataset's signal is weak to start with, our methods may incorrectly amplify errors or biases.  The application of our techniques in practical settings should be evaluated before deployment.
This work experimented with publicly available datasets which require no additional annotation from humans. 

\section*{Acknowledgement}
The author thanks the anonymous reviewers for their valuable advice and Taha Aksu for his diligent editing of the manuscript.
\bibliography{DST}
\bibliographystyle{acl_natbib}

\end{document}